%% file: main.tex
\documentclass[10pt]{article}
\usepackage[preprint]{tmlr}

\input{tex/preamble.tex}

\newcommand{\SynthTemplate}{Template v1}
\newcommand{\SynthLLM}{LLM augmentation}

\newif\ifblinded
\blindedfalse

\title{LLM Readiness Harness: Evaluation, Observability, and CI Gates for LLM/RAG Applications}
\author{
\name Alexandre Cristov\~ao Maiorano \email alexandre@lumytics.com \\
\addr Lumytics
}

\def\month{03}
\def\year{2026}
\def\openreview{}

\newcommand{\DisclosureSectionTitle}{Acknowledgments and AI Tools Disclosure}
\newcommand{\ReplicationAvailabilityText}{A companion public reproducibility repository, \url{https://github.com/alemaiorano/llm-readiness-harness-reproducibility}, provides manuscript sources, frozen analysis outputs, lightweight export scripts, and claim-to-artifact traceability. The public artifact is intentionally narrower than the full internal experimentation stack and is designed for independent verification of the released manuscript.}

\begin{document}
\input{paper/main_shared.tex}
\end{document}

%% file: tex/preamble.tex
\usepackage[T1]{fontenc}
\usepackage{lmodern}
\usepackage{amsmath}
\usepackage{booktabs}
\usepackage{array}
\usepackage{graphicx}
\usepackage[hypertexnames=false]{hyperref}
\usepackage[expansion=false]{microtype}
\usepackage{tikz}
\usepackage{pgfplots}
\usepackage{tcolorbox}
\usepackage{float}
\usepackage{placeins}
\usepackage{needspace}
\usepackage{xurl}
\usepackage{pdflscape}

\usetikzlibrary{arrows.meta,positioning,fit,backgrounds}
\pgfplotsset{compat=1.18}
\tcbuselibrary{listings,skins,breakable}

\newtcblisting{codebox}{
  listing only,
  breakable,
  colback=gray!5,
  colframe=gray!60,
  boxrule=0.4pt,
  left=6pt,right=6pt,top=4pt,bottom=4pt,
  listing options={basicstyle=\ttfamily\footnotesize,breaklines=true,columns=fullflexible}
}

%% file: paper/main_shared.tex
\maketitle

\begin{abstract}
We present a readiness harness for LLM and RAG applications that turns evaluation into a
deployment decision workflow. The system combines automated benchmarks, OpenTelemetry
observability, and CI quality gates under a minimal API contract, then aggregates workflow
success, policy compliance, groundedness, retrieval hit rate, cost, and p95 latency into
scenario-weighted readiness scores with Pareto frontiers. We evaluate the harness on ticket-routing
workflows and BEIR grounding tasks (SciFact and FiQA) with full Azure matrix coverage
(162/162 valid cells across datasets, scenarios, retrieval depths, seeds, and models). Results
show that readiness is not a single metric: on FiQA under \texttt{sla-first} at k=5,
\texttt{gpt-4.1-mini} leads in readiness and faithfulness, while \texttt{gpt-5.2} pays a
substantial latency cost; on SciFact, models are closer in quality but still separable
operationally. Ticket-routing regression gates consistently reject unsafe prompt variants,
demonstrating that the harness can block risky releases instead of merely reporting offline
scores. The result is a reproducible, operationally grounded framework for deciding whether an
LLM or RAG system is ready to ship.
\end{abstract}

\section{Introduction}
\input{paper/sections/01-introduction}

\section{System Overview}
\input{paper/sections/02-system}

\section{Readiness Scoring}
\input{paper/sections/03-method}

\section{Benchmarks and Datasets}
\input{paper/sections/04-benchmark}

\section{Experiments}
\input{paper/sections/05-experiments}

\section{Validation and Regression Tests}
\input{paper/sections/06-validation}

\section{Lessons Learned}
\input{paper/sections/07-lessons}

\section{Limitations and Threats to Validity}
\input{paper/sections/08-limitations}

\section{Related Work}
\input{paper/sections/09-related}

\section{Conclusion}
\input{paper/sections/10-conclusion}

\appendix
\raggedbottom
\section{Additional Tables}
\input{paper/tables/summary_table.tex}
\input{paper/tables/fiqa_summary_table.tex}
\input{paper/tables/weight_ablation_table.tex}
\input{paper/tables/ticket_summary.tex}
\input{paper/tables/ticket_seed_stats.tex}

\section{Prompts, Schemas, and Pseudocode}
\input{paper/sections/appendix-extras}

\section{Ticket Routing Results (Full)}
\label{app:ticket-results-full}
\input{paper/tables/ticket_summary_full.tex}

\section{Ticket Error Summary}
\label{app:ticket-error-full}
\input{paper/tables/ticket_error_summary.tex}
\input{paper/tables/ticket_error_full.tex}

\section*{\DisclosureSectionTitle}
This research leveraged AI-assisted tools to support code development and manuscript preparation, under full human oversight and accountability. The following tools were used:
\begin{itemize}
    \item \textbf{Language models:} OpenAI Codex CLI (a code-assistance tool powered by the GPT-5 family, distinct from the deprecated Codex model series), Anthropic Claude Code (Opus/Sonnet), and Google Gemini CLI were used for code drafting/review and language refinement of author-written text.
    \item \textbf{Web search:} MCP Tavily integration was used to support literature discovery and fact-checking.
\end{itemize}
All scientific arguments, empirical methodology, statistical analysis, research questions, and conclusions were conceived, developed, and validated by the authors. AI assistance was limited to code drafting/review and language refinement; it was not used to generate or alter experimental data, metrics, or conclusions. Authors reviewed and validated all AI-assisted outputs before inclusion. AI systems were not listed as authors. No generative-AI artwork or figures were used in the manuscript. \ReplicationAvailabilityText

{\sloppy\Urlmuskip=0mu plus 2mu\relax
\bibliographystyle{tmlr}
\bibliography{references}
}

%% file: paper/sections/01-introduction.tex
Production teams struggle to apply release-engineering rigor to LLM and RAG systems.
Evaluation, observability, and quality gates are often ad hoc, leading to silent regressions,
unpredictable cost, and policy violations. We target the hypothesis that higher textual quality
does not necessarily maximize utility once cost, latency, groundedness, and policy constraints
are considered.

We propose a readiness harness that combines automated evaluation, observability, and CI gates,
then surfaces cost--utility frontiers for deployment decisions. The harness is deliberately
pragmatic: it defines a stable API contract, produces run-level artifacts, and enables
scenario-specific scoring for cost-first, risk-first, and SLA-first rollouts.
While prior tools already measure RAG quality, we focus on release readiness by binding those
metrics to observability, CI gates, and operational constraints.
This broader framing also creates an operational substrate for follow-on work on automated
self-testing quality gates. In a later companion case study, automated self-testing is
specialized into a longitudinal PROMOTE/HOLD/ROLLBACK release workflow for a deployed
multi-agent application \citep{maiorano2026selftesting}. We therefore position the present
harness as a general readiness layer rather than a dogfooding-only protocol.

\paragraph{Contributions.}
\begin{itemize}
  \item An end-to-end readiness harness architecture (batch runner + API + CI gates).
  \item A multi-dimensional readiness score and cost-utility frontier analysis.
  \item A benchmark plan for workflow tickets (T1/T2) and retrieval (T3/BEIR).
\end{itemize}

%% file: paper/sections/02-system.tex
The system consists of a local API service for inference, a batch worker for benchmarking,
OpenTelemetry instrumentation \citep{opentelemetry-collector,opentelemetry-docs}, and CI quality
gates via promptfoo \citep{promptfoo-ci,promptfoo-http}. The API exposes a stable contract for
single-case inference and batch runs, returning structured outputs plus metadata for cost,
latency, and trace correlation. The batch worker orchestrates dataset sampling, stores artifacts
per run (reports, scorecards, and frontiers), and can be triggered in CI or nightly schedules.

The observability layer creates spans for routing, retrieval, generation, and policy validation,
enabling post-hoc debugging of regressions. CI gates are enforced via promptfoo to detect
schema violations, policy failures, and quality regressions before deployment.

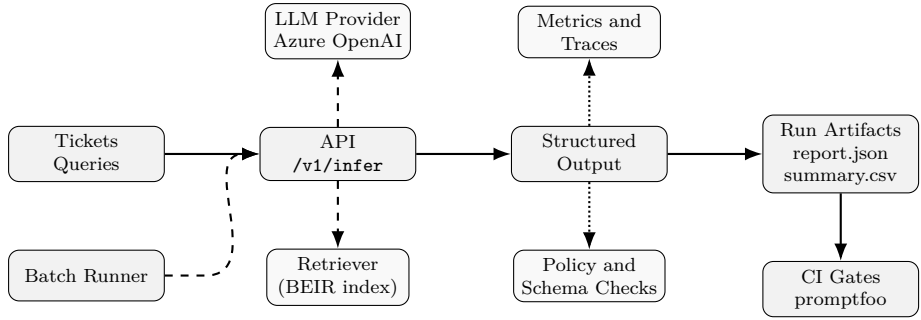
\begin{figure}[t]
  \centering
  \begin{tikzpicture}[
    node distance=1.0cm and 1.4cm,
    block/.style={draw, rounded corners, align=center, minimum width=2.3cm, minimum height=0.75cm, fill=gray!10},
    sub/.style={draw, rounded corners, align=center, minimum width=2.1cm, minimum height=0.7cm, fill=gray!5},
    line/.style={-{Latex}, thick},
    call/.style={-{Latex}, thick, dashed},
    aux/.style={-{Latex}, thick, densely dotted},
    font=\footnotesize,
    scale=0.9,
    transform shape
  ]
    \node[block] (input) {Tickets\\Queries};
    \node[block, right=of input] (api) {API\\\texttt{/v1/infer}};
    \node[block, right=of api] (output) {Structured\\Output};
    \node[sub, below=of api] (retriever) {Retriever\\(BEIR index)};
    \node[sub, above=of api] (llm) {LLM Provider\\Azure OpenAI};
    \node[sub, below=of output] (checks) {Policy and\\Schema Checks};
    \node[sub, above=of output] (metrics) {Metrics and\\Traces};
    \node[block, right=of output] (artifacts) {Run Artifacts\\report.json\\summary.csv};
    \node[block, below=of artifacts] (ci) {CI Gates\\promptfoo};
    \node[block, below=of input] (batch) {Batch Runner};

    \draw[line] (input) -- (api);
    \draw[line] (api) -- (output);
    \draw[call] (api) -- (retriever);
    \draw[call] (api) -- (llm);
    \draw[aux] (output) -- (checks);
    \draw[aux] (output) -- (metrics);
    \draw[line] (output) -- (artifacts);
    \draw[line] (artifacts) -- (ci);
    \draw[call] (batch.east) -- ++(0.6cm,0) to[out=0,in=180,looseness=1.2] (api.west);
  \end{tikzpicture}
  \caption{System architecture and instrumentation points.}
  \label{fig:architecture}
\end{figure}

The API contract and instrumentation schema standardize required fields (run id, dataset id,
pipeline metadata, latency, token usage, and cost estimates) and optional trace ids for
cross-system correlation. For privacy and governance, traces record hashed ticket identifiers
and can redact or exclude raw user content, with retention and access controlled at the
artifact storage layer.
Figure~\ref{fig:architecture} summarizes the runtime components and instrumentation points
captured by the harness.

Table~\ref{tab:otel-schema} summarizes the default span names and attributes captured by the
instrumentation layer for each inference run.
\begin{table}[t]
\centering
\caption{Default OpenTelemetry spans and attributes.}
\label{tab:otel-schema}
\small
\setlength{\tabcolsep}{4pt}
\resizebox{\linewidth}{!}{%
\begin{tabular}{l >{\raggedright\arraybackslash}p{0.28\linewidth} >{\raggedright\arraybackslash}p{0.55\linewidth}}
\toprule
Span & Purpose & Key attributes \\
\midrule
infer & Root span for the inference request & run\_id, dataset\_id, pipeline\_version, scenario, latency\_ms, tokens\_in, tokens\_out, cost\_usd\_est \\
route.classify & Route label prediction & ticket\_id (hashed), dataset\_id, predicted\_label, confidence, latency\_ms \\
respond.finalize & Escalation decision & ticket\_id (hashed), should\_escalate, latency\_ms \\
rag.retrieve & Retrieval step & top\_k, index\_id, retrieved\_doc\_ids, latency\_ms \\
rag.generate & LLM generation & model\_id, provider, cache\_hit, latency\_ms \\
validate.policy & Policy checks & pass, violation\_types, latency\_ms \\
\bottomrule
\end{tabular}
}
\end{table}

%% file: paper/sections/03-method.tex
We compute readiness using a small set of observable dimensions. Each run yields
metrics for workflow success, policy compliance, groundedness (faithfulness),
retrieval hit@k (BEIR), cost per task, and p95 latency \citep{beir2021,ragas2023}.
We define a weighted readiness score
$R = \sum_i w_i \cdot m_i$, where weights are scenario-specific:
cost-first, risk-first, and SLA-first. To preserve auditability in the presence
of missing metrics, we compute
$R = \frac{\sum_{i \in \mathcal{P}} w_i m_i}{\sum_{i \in \mathcal{P}} w_i}$,
where $\mathcal{P}$ are the metrics present in a run. Missing metrics are explicitly
reported alongside the score.

All scalar metrics are normalized to the $[0,1]$ range. Rate-based signals (e.g.,
faithfulness, retrieval hit@k, policy pass rates) are used directly; cost and
latency are reported with their raw values and mapped to optional normalized
scores when budgets or thresholds are configured. Policy compliance is also
treated as a hard gate in CI: runs that violate policies cannot pass gates even
if their scalar score is high. We also compute cost-utility frontiers by
identifying non-dominated pipelines across cost and quality dimensions.

Scenario weights (sum to 1) are fixed and published in code. Table~\ref{tab:scenario-weights} presents the weights used for each evaluation dimension across the three deployment scenarios.
\begin{table}[h]
\centering
\caption{Scenario weights used in readiness scoring.}
\label{tab:scenario-weights}
\small
\begin{tabular}{lrrrrrr}
\toprule
\textbf{Scenario} & \textbf{Workflow} & \textbf{Policy} & \textbf{Faithfulness} & \textbf{Retrieval} & \textbf{Cost} & \textbf{SLA} \\
\midrule
cost-first & 0.20 & 0.20 & 0.15 & 0.15 & 0.20 & 0.10 \\
risk-first & 0.15 & 0.25 & 0.20 & 0.15 & 0.10 & 0.15 \\
SLA-first & 0.20 & 0.15 & 0.15 & 0.10 & 0.10 & 0.30 \\
\bottomrule
\end{tabular}
\end{table}

Cost is estimated from model token usage using published API pricing
\citep{openai-api-pricing}. The scoring function is intentionally simple to preserve
auditability, enable calibration by product owners, and isolate the impact of
ablations on each dimension. A weight ablation varying scenario assignments is
reported in Table~\ref{tab:weight-ablation} (Appendix).

Not all dimensions are populated for every task type. For BEIR retrieval tasks
(T3), workflow success and policy compliance are not applicable and are excluded
from $\mathcal{P}$; the score is thus driven by faithfulness, retrieval hit@k,
cost, and SLA. For ticket routing tasks (T1/T2), retrieval hit@k is not
applicable; the score is driven by workflow success, policy compliance,
faithfulness, cost, and SLA. This per-task scoping is explicit in exported
artifacts so that cross-task score comparisons are made with awareness of
which dimensions are active.

Pareto frontiers are computed per scenario using available metrics, maximizing
quality signals (e.g., faithfulness, retrieval hit@k) while minimizing cost and
latency. We report the chosen metric set for each run to make dominance checks
explicit and reproducible.

%% file: paper/sections/04-benchmark.tex
We benchmark two task families, consisting of three defined tasks. Task 1 (T1) and Task 2 (T2) cover customer-support workflows using public
ticket datasets to evaluate routing (T1), escalation decisions, and policy checks (T2).
Task 3 (T3) evaluates retrieval grounding using BEIR subsets (scifact for factual grounding
and fiqa for domain shift) \citep{beir2021}. This separation keeps the framing consistent
with both product needs (workflow and policy) and scientific rigor (retrieval OOD
generalization).

\paragraph{Datasets and splits.}
For T1/T2, we use IT Service Ticket Classification (Kaggle) \citep{it-ticket-kaggle}
and Multilingual Customer Support Tickets (Kaggle/HF) \citep{multilingual-ticket-kaggle}
to cover structured routing and multilingual stress cases. We generate stratified
80/10/10 train/val/test splits with a fixed seed and publish a frozen regression
subset (from the test split) for CI gates. Each split and regression suite is
versioned with a manifest that records the seed, counts, and source URI.
The multilingual dataset contains 28,587 tickets (en=16,338; de=12,249) across
10 queues, with median ticket length 426 characters.
The IT service dataset contains 47,837 tickets across 8 queues (Hardware, HR Support,
Access, Miscellaneous, Storage, Purchase, Internal Project, Administrative rights),
giving a clean routing baseline and enabling TF-IDF versus LLM comparisons under a
stable label schema. The multilingual dataset stresses language robustness (English
and German) and queue imbalance, mirroring real enterprise support distributions.
Appendix~\ref{app:dataset-examples} includes short examples of each dataset
family for concreteness.

We include a minimal regression set with ambiguous tickets and prompt-injection-style
instructions to stress policy and grounding. For each dataset, we sample a
fixed number of queries and run controlled ablations over retrieval depth and
prompting configurations.

For T3, we use BEIR SciFact and FiQA \citep{beir2021}. SciFact consists of scientific
claims paired with evidence-bearing abstracts, emphasizing factual grounding.
FiQA covers finance questions with supporting passages from a domain-specific
corpus, representing domain shift. We keep the original BEIR query/corpus splits
and evaluate retrieval hit@k independently from workflow metrics.

\paragraph{Synthetic dataset generation (product).}
To support product usage and avoid non-commercial license constraints, we define
a synthetic ticket generator with audit-ready guardrails. We use templates as a
safe base and optionally apply self-instruct style LLM expansion for diversity
\citep{self-instruct,unnatural-instructions,eda}. Labels (queue/priority/escalation)
are fixed by rules and validated post-hoc, and distributions are enforced via
stratified quotas. We filter with schema and policy checks, deduplication, and
similarity thresholds, and publish manifests and dataset cards for traceability
\citep{datasheets,datastatements}.
The synthetic suite is designed to be multilingual (en/pt/es) to mirror
enterprise support coverage, and is intended for product regression and CI gates
rather than replacing real-world evaluation datasets.
Table~\ref{tab:synthetic-comparison} summarizes template versus LLM generation
quality metrics for the synthetic ticket suite.
The LLM comparison uses a 500-sample run due to local compute limits, while the
template run uses 4,000 samples for stable distribution checks.

\input{paper/tables/synthetic_comparison_table.tex}

In this synthetic track, both generators remain schema/policy-safe (zero schema
errors and policy violations), but they differ in distribution behavior. The
template baseline reaches perfect quota compliance (quality score 1.0000 and
zero distribution violations), while the LLM-augmentation run is close in
aggregate quality (0.9936) with higher lexical diversity and novelty
(Lex 0.0898; Uniq 0.3960), but incurs one distribution violation at n=500.
We treat this as operational guidance for synthetic-data generation behavior, not
a strict ranking, because sample sizes and compute constraints differ across
tracks.

\paragraph{Task definitions.}
T1 success is correct routing (route\_label) against dataset labels. T2 success
requires zero policy violations and correct escalation when a gold escalation
label is available; otherwise we report escalation rate separately and treat
policy as the hard gate. Policy rules include a minimal credential-collection
ban (e.g., asks\_for\_password), treated as a hard gate for CI. For T3, we use BEIR
SciFact and FiQA, and report retrieval metrics separately from end-to-end workflow
outcomes to avoid conflating retrieval relevance with task readiness.

%% file: paper/tables/synthetic_comparison_table.tex
\begin{table}[t]
\centering
\small
\caption{Synthetic dataset quality comparison (template vs LLM).}
\label{tab:synthetic-comparison}
\setlength{\tabcolsep}{3pt}
\renewcommand{\arraystretch}{1.05}
\begin{tabular}{lrrrrrrrrrr}
\toprule
Dataset & $n$ & Q & JS & Lex & Uniq & Esc & Sch & Pol & Dup & Viol \\
\midrule
\SynthTemplate & 4000 & 1.0000 & 0.0000 & 0.0036 & 0.0695 & 0.1923 & 0.0000 & 0.0000 & 0.0000 & 0 \\
\SynthLLM & 500 & 0.9936 & 0.0064 & 0.0898 & 0.3960 & 0.2040 & 0.0000 & 0.0000 & 0.0000 & 1 \\
\bottomrule
\end{tabular}
\end{table}

%% file: paper/sections/05-experiments.tex
Baselines include prompt-only LLM routing and RAG with policy gates.
This version reports the hosted Azure matrix end-to-end (OpenAI model deployments:
\texttt{gpt-4.1}, \texttt{gpt-4.1-mini}, \texttt{gpt-5.2}) alongside a
TF-IDF + LinearSVC baseline for ticket routing tasks.

For BEIR, we execute the full matrix on SciFact and FiQA:
3 scenarios (\texttt{cost/risk/sla-first}) $\times$ 3 top-k values (3/5/10)
$\times$ 3 seeds (42/1337/2024) $\times$ 3 models, with \texttt{sample\_n=60}
per run. After filtering to latest valid runs per
(dataset, scenario, k, seed, model), coverage is complete:
27/27 per model in each dataset (81 SciFact + 81 FiQA clean runs).
Retrieval ablation uses \texttt{sample\_n=80} (\texttt{sla-first}, k=5, seed=2024).
Ticket sweeps run on two datasets with prompt v1/v2, three seeds, and
\texttt{sample\_n=200}; ticket regressions (baseline/bias/policy) use
\texttt{sample\_n=60}. Azure execution uses worker parallelism and resume/checkpoint
support to recover safely from connection interruptions.

We report cost, latency, workflow success, policy pass/violations, faithfulness,
and retrieval hit@k. Answer relevance (a RAGAS sub-metric) was collected but
returned empty across all runs under the current RAGAS configuration, so it is
excluded from aggregate scores; the score formula uses only present metrics
via the $\mathcal{P}$ mechanism (Section 3). Artifacts are exported to CSV/JSON/LaTeX
from the same run manifests (Appendix~\ref{app:artifact-examples}).

The BEIR core results (Table~\ref{tab:seed-stats}) show stable model separation.
At FiQA, k=5, \texttt{sla-first}, \texttt{gpt-4.1-mini} leads on readiness
(0.808 $\pm$ 0.022) and faithfulness (0.777 $\pm$ 0.069), followed by
\texttt{gpt-4.1} (0.782 $\pm$ 0.016; faithfulness 0.665 $\pm$ 0.047), while
\texttt{gpt-5.2} is lower on readiness and faithfulness (0.710 $\pm$ 0.014;
0.387 $\pm$ 0.016) and substantially slower (p95 $\sim$9.0s vs 3.6--4.0s).
At SciFact, k=5, \texttt{sla-first}, the gap is smaller:
\texttt{gpt-4.1-mini} 0.811 $\pm$ 0.017,
\texttt{gpt-4.1} 0.798 $\pm$ 0.033, and
\texttt{gpt-5.2} 0.798 $\pm$ 0.010; latency remains materially higher for
\texttt{gpt-5.2} (p95 $\sim$6.2s).
Per-run sample metrics for SciFact and FiQA are provided in
Tables~\ref{tab:summary} and \ref{tab:fiqa-summary} in the Appendix.

Across models, increasing k improves retrieval hit@k (SciFact: 0.8167 $\rightarrow$
0.8889; FiQA: 0.5278 $\rightarrow$ 0.6444 from k=3 to k=10) but increases cost
($\sim$0.007--0.008 to $\sim$0.020 per run) and does not always improve
readiness under cost/SLA-weighted scoring. Pareto frontiers and non-dominated
runs are shown in Figures~\ref{fig:frontier-scifact} (SciFact) and
\ref{fig:frontier-fiqa} (FiQA); the non-dominated runs for SciFact are
enumerated in Table~\ref{tab:pareto}. Tables keep explicit model labels to
avoid cross-model ambiguity.
\IfFileExists{paper/tables/pareto_table.tex}{\input{paper/tables/pareto_table.tex}}{}
\IfFileExists{paper/figures/pareto_frontier_scifact.tex}{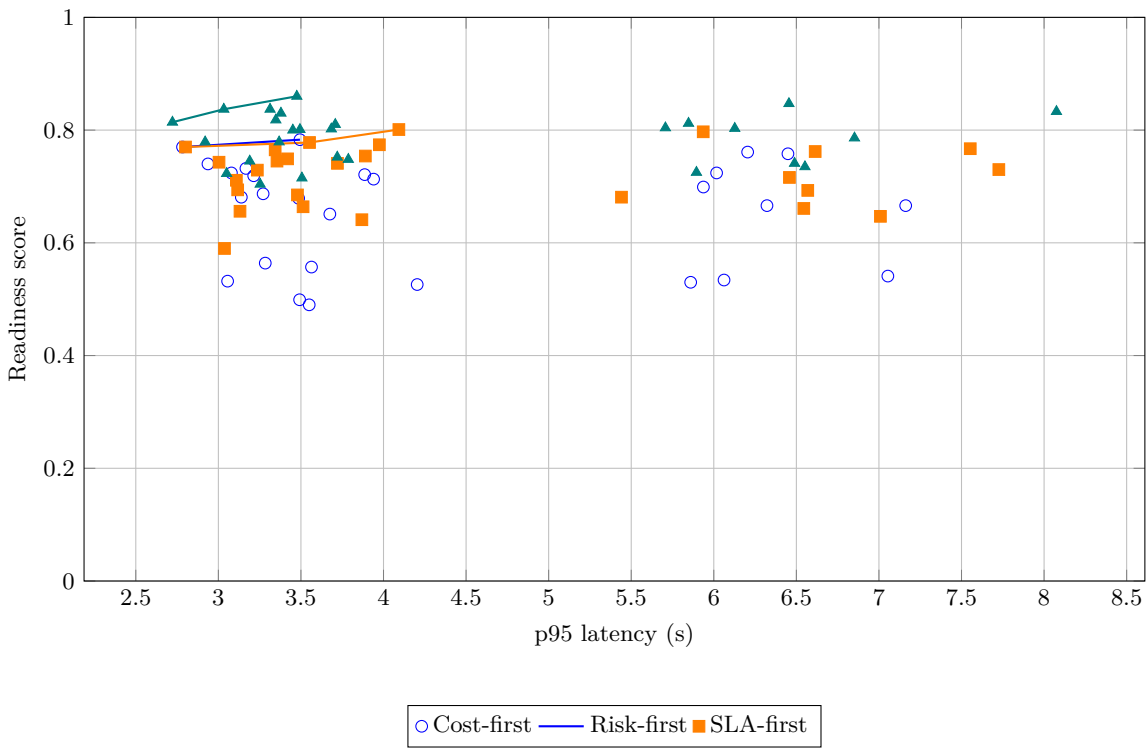
\begin{figure}[t]\centering\input{paper/figures/pareto_frontier_scifact.tex}\caption{Latency-utility frontiers across scenarios (SciFact; p95 latency vs readiness score).}\label{fig:frontier-scifact}\end{figure}}{}
\IfFileExists{paper/figures/pareto_frontier_fiqa.tex}{\begin{figure}[t]\centering\input{paper/figures/pareto_frontier_fiqa.tex}\caption{Latency-utility frontiers across scenarios (FiQA; p95 latency vs readiness score).}\label{fig:frontier-fiqa}\end{figure}
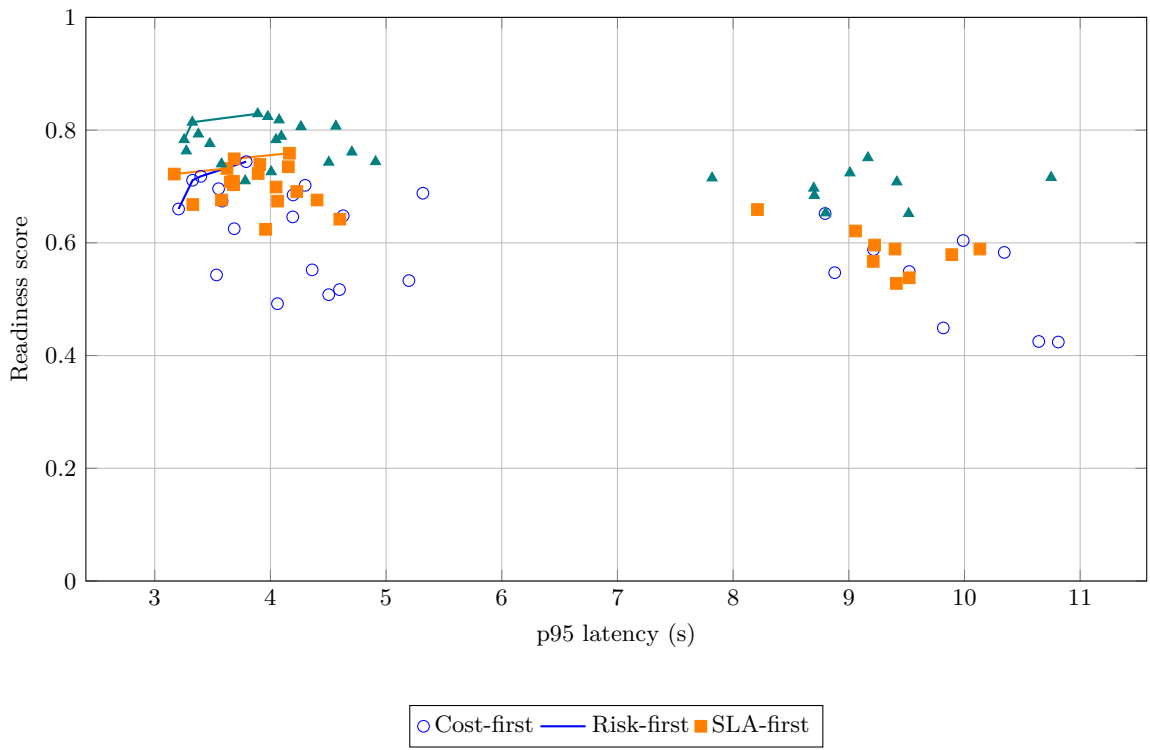}{}

Retriever ablation on SciFact (Table~\ref{tab:retriever-ablation}) shows BM25
without reranker maximizes hit@k (0.850), while BM25+reranker gives the best
faithfulness in this slice (0.525) with the lowest p95 latency (2859 ms).
Note that this ablation covers a single dataset, seed, and k value (SciFact,
seed=2024, k=5); broader generalization should be validated across seeds and
datasets before drawing definitive retriever conclusions.
\IfFileExists{paper/tables/retriever_ablation_table.tex}{\input{paper/tables/retriever_ablation_table.tex}}{}

For ticket routing, compact and full summaries are reported in
Tables~\ref{tab:ticket-results} and \ref{tab:ticket-results-full}.
On customer-support tickets (10 queues; random baseline $\approx$10\%),
prompt v1/v2 are close in routing accuracy (0.295 vs 0.292 mean), with v2
reducing p95 latency (3150 ms vs 3444 ms) at slightly lower workflow success
(0.148 vs 0.167).
On IT tickets (8 queues; random baseline $\approx$12.5\%), v2 improves routing
(0.408 vs 0.393 mean accuracy) and p95 latency (2447 ms vs 2743 ms), while
workflow success remains close (0.358 vs 0.362). Routing accuracy in both
domains is above the random baseline but reflects the inherent complexity of
multi-queue classification under a zero-shot prompt contract.
Seed robustness for these runs is summarized in Table~\ref{tab:ticket-seed-stats}.
Detailed per-run values and regression variants are included in
Table~\ref{tab:ticket-results-full} (Appendix~\ref{app:ticket-results-full}).

We run each configuration with seeds 42/1337/2024 and report mean $\pm$ std
(Table~\ref{tab:seed-stats}; Figure~\ref{fig:seed-robustness}). p95 latency is computed from
per-sample run logs and summarized in the released analysis outputs. Azure runs support larger samples with lower
tail latency and full-model matrix coverage, while preserving artifact-level
reproducibility.
\IfFileExists{paper/tables/seed_stats_table.tex}{\input{paper/tables/seed_stats_table.tex}}{}
\IfFileExists{paper/figures/seed_robustness.tex}{\begin{figure}[t]\centering\input{paper/figures/seed_robustness.tex}\caption{Seed robustness for readiness score (mean $\pm$ std; top\_k=5).}\label{fig:seed-robustness}\end{figure}
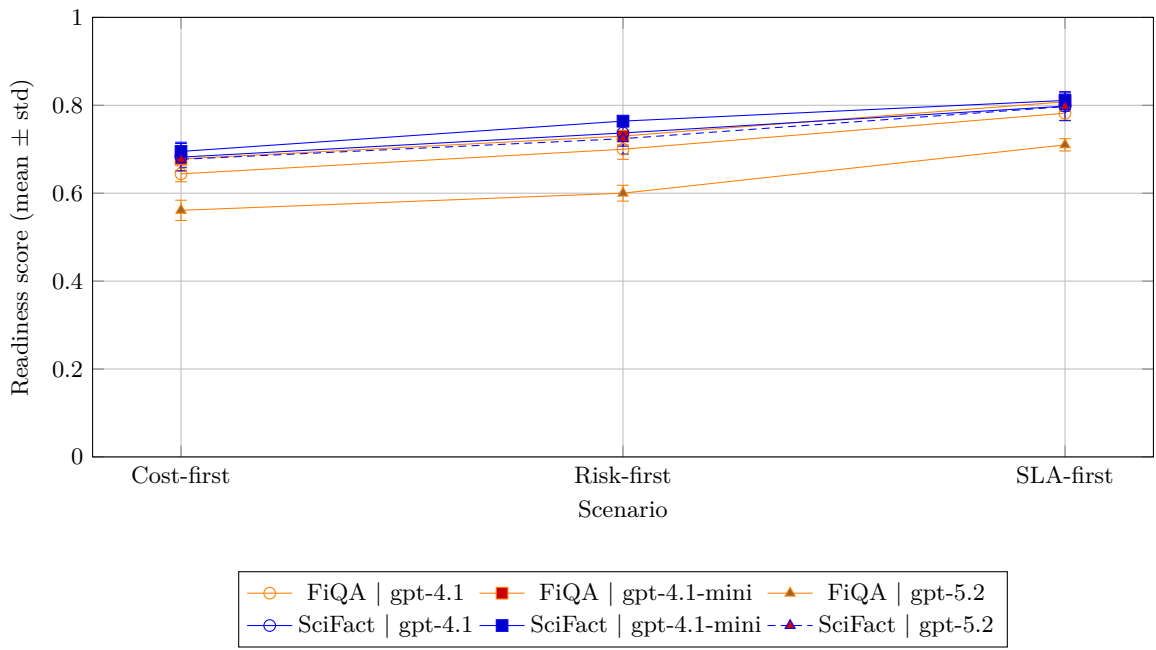}{}

For analysis tables, we filter to latest valid runs per
(dataset, scenario, k, seed, model, provider), requiring
\texttt{ragas\_status=ok}, zero evaluator errors, and consistent dataset/pipeline/sample
settings. The resulting filtered analysis sets are released with the public reproducibility artifact
for SciFact and FiQA.

%% file: paper/tables/pareto_table.tex
\begin{table}[!ht]
\centering
\caption{Pareto frontier (all scenarios; dataset: SciFact).}
\label{tab:pareto}
\small
\setlength{\tabcolsep}{4pt}
\resizebox{\linewidth}{!}{%
\begin{tabular}{l l r r r r r r r}
\toprule
Scenario & Model & k & Seed & Score & Faithfulness & Hit@k & Cost (\$) & p95 latency (ms) \\
\midrule
Risk-first & gpt-4.1-mini & 10 & 1337 & 0.664 & 0.567 & 0.900 & 0.0200 & 3513 \\
Risk-first & gpt-4.1-mini & 10 & 42 & 0.685 & 0.644 & 0.883 & 0.0205 & 3479 \\
Risk-first & gpt-4.1-mini & 3 & 1337 & 0.741 & 0.539 & 0.783 & 0.0075 & 3721 \\
Risk-first & gpt-4.1-mini & 3 & 2024 & 0.749 & 0.515 & 0.850 & 0.0076 & 3419 \\
Risk-first & gpt-4.1-mini & 3 & 42 & 0.801 & 0.700 & 0.817 & 0.0077 & 4093 \\
Risk-first & gpt-4.1-mini & 5 & 1337 & 0.765 & 0.674 & 0.833 & 0.0111 & 3344 \\
Risk-first & gpt-4.1-mini & 5 & 2024 & 0.754 & 0.632 & 0.850 & 0.0113 & 3890 \\
Risk-first & gpt-4.1-mini & 5 & 42 & 0.774 & 0.707 & 0.833 & 0.0114 & 3975 \\
Cost-first & gpt-4.1 & 10 & 1337 & 0.532 & 0.563 & 0.900 & 0.0201 & 3056 \\
Cost-first & gpt-4.1 & 10 & 42 & 0.557 & 0.676 & 0.883 & 0.0205 & 3564 \\
Cost-first & gpt-4.1 & 3 & 1337 & 0.724 & 0.526 & 0.783 & 0.0076 & 3079 \\
Cost-first & gpt-4.1 & 3 & 2024 & 0.740 & 0.530 & 0.850 & 0.0077 & 2937 \\
Cost-first & gpt-4.1 & 3 & 42 & 0.770 & 0.687 & 0.817 & 0.0077 & 2783 \\
Cost-first & gpt-4.1 & 5 & 1337 & 0.681 & 0.571 & 0.833 & 0.0112 & 3139 \\
Risk-first & gpt-4.1 & 10 & 1337 & 0.656 & 0.542 & 0.900 & 0.0201 & 3131 \\
Risk-first & gpt-4.1 & 10 & 2024 & 0.590 & 0.358 & 0.883 & 0.0203 & 3037 \\
Risk-first & gpt-4.1 & 10 & 42 & 0.694 & 0.669 & 0.883 & 0.0205 & 3117 \\
Risk-first & gpt-4.1 & 5 & 1337 & 0.729 & 0.567 & 0.833 & 0.0112 & 3236 \\
Risk-first & gpt-4.1 & 5 & 42 & 0.770 & 0.696 & 0.833 & 0.0114 & 2801 \\
SLA-first & gpt-4.1 & 10 & 1337 & 0.723 & 0.534 & 0.900 & 0.0201 & 3050 \\
SLA-first & gpt-4.1 & 10 & 42 & 0.745 & 0.638 & 0.883 & 0.0205 & 3190 \\
SLA-first & gpt-4.1 & 3 & 2024 & 0.814 & 0.501 & 0.850 & 0.0076 & 2722 \\
SLA-first & gpt-4.1 & 3 & 42 & 0.837 & 0.628 & 0.817 & 0.0077 & 3033 \\
SLA-first & gpt-4.1 & 5 & 42 & 0.837 & 0.752 & 0.833 & 0.0114 & 3313 \\
Cost-first & gpt-4.1-mini & 10 & 42 & 0.564 & 0.707 & 0.883 & 0.0205 & 3284 \\
Cost-first & gpt-4.1-mini & 3 & 1337 & 0.732 & 0.551 & 0.783 & 0.0075 & 3167 \\
Cost-first & gpt-4.1-mini & 3 & 2024 & 0.721 & 0.446 & 0.850 & 0.0076 & 3886 \\
Cost-first & gpt-4.1-mini & 3 & 42 & 0.783 & 0.739 & 0.817 & 0.0077 & 3494 \\
Cost-first & gpt-4.1-mini & 5 & 1337 & 0.687 & 0.588 & 0.833 & 0.0111 & 3271 \\
Cost-first & gpt-4.1-mini & 5 & 42 & 0.719 & 0.740 & 0.833 & 0.0114 & 3214 \\
SLA-first & gpt-4.1-mini & 10 & 1337 & 0.748 & 0.641 & 0.900 & 0.0200 & 3787 \\
SLA-first & gpt-4.1-mini & 10 & 42 & 0.752 & 0.669 & 0.883 & 0.0205 & 3720 \\
SLA-first & gpt-4.1-mini & 3 & 1337 & 0.818 & 0.559 & 0.783 & 0.0075 & 3348 \\
SLA-first & gpt-4.1-mini & 3 & 2024 & 0.801 & 0.445 & 0.850 & 0.0076 & 3494 \\
SLA-first & gpt-4.1-mini & 3 & 42 & 0.860 & 0.727 & 0.817 & 0.0077 & 3475 \\
SLA-first & gpt-4.1-mini & 5 & 1337 & 0.802 & 0.591 & 0.833 & 0.0111 & 3684 \\
SLA-first & gpt-4.1-mini & 5 & 2024 & 0.800 & 0.577 & 0.850 & 0.0112 & 3451 \\
SLA-first & gpt-4.1-mini & 5 & 42 & 0.830 & 0.724 & 0.833 & 0.0114 & 3379 \\
Risk-first & gpt-5.2 & 3 & 2024 & 0.767 & 0.579 & 0.850 & 0.0080 & 7553 \\
Cost-first & gpt-5.2 & 3 & 2024 & 0.758 & 0.626 & 0.850 & 0.0080 & 6449 \\
SLA-first & gpt-5.2 & 3 & 2024 & 0.833 & 0.597 & 0.850 & 0.0080 & 8075 \\
SLA-first & gpt-5.2 & 5 & 2024 & 0.804 & 0.605 & 0.850 & 0.0116 & 5707 \\
\bottomrule
\end{tabular}
}
\end{table}

%% file: paper/figures/pareto_frontier_scifact.tex
\begin{tikzpicture}
\begin{axis}[
width=0.95\linewidth,
height=0.55\linewidth,
xlabel={p95 latency (s)},
ylabel={Readiness score},
ymin=0,
ymax=1,
grid=major,
legend style={font=\small, at={(0.5,-0.22)}, anchor=north, legend columns=3},
tick label style={font=\small},
label style={font=\small},
]
\addplot[only marks, mark=o, mark size=2.2pt, color=blue] coordinates {(3.056,0.532) (3.550,0.490) (3.564,0.557) (3.079,0.724) (2.937,0.740) (2.783,0.770) (3.139,0.681) (3.675,0.651) (3.940,0.713) (4.204,0.526) (3.492,0.499) (3.284,0.564) (3.167,0.732) (3.886,0.721) (3.494,0.783) (3.271,0.687) (3.486,0.679) (3.214,0.719) (5.860,0.530) (6.062,0.534) (7.054,0.541) (6.017,0.724) (6.449,0.758) (6.205,0.761) (7.162,0.666) (6.322,0.666) (5.937,0.699)};
\addlegendentry{Cost-first}
\addplot[thick, color=blue, mark=none] coordinates {(2.783,0.770) (3.494,0.783)};
\addplot[only marks, mark=square*, mark size=2.2pt, color=orange] coordinates {(3.513,0.664) (3.869,0.641) (3.479,0.685) (3.721,0.741) (3.419,0.749) (4.093,0.801) (3.344,0.765) (3.890,0.754) (3.975,0.774) (3.131,0.656) (3.037,0.590) (3.117,0.694) (3.355,0.745) (3.004,0.743) (3.552,0.778) (3.236,0.729) (3.109,0.711) (2.801,0.770) (7.009,0.647) (6.545,0.661) (5.441,0.681) (7.726,0.730) (7.553,0.767) (5.936,0.797) (6.569,0.693) (6.458,0.716) (6.614,0.762)};
\addlegendentry{Risk-first}
\addplot[thick, color=orange, mark=none] coordinates {(2.801,0.770) (3.552,0.778) (4.093,0.801)};
\addplot[only marks, mark=triangle*, mark size=2.2pt, color=teal] coordinates {(3.050,0.723) (3.253,0.704) (3.190,0.745) (3.708,0.810) (2.722,0.814) (3.033,0.837) (2.920,0.779) (3.368,0.779) (3.313,0.837) (3.787,0.748) (3.506,0.715) (3.720,0.752) (3.348,0.818) (3.494,0.801) (3.475,0.860) (3.684,0.802) (3.451,0.800) (3.379,0.830) (5.895,0.725) (6.552,0.735) (6.489,0.741) (5.847,0.812) (8.075,0.833) (6.455,0.847) (6.852,0.786) (5.707,0.804) (6.127,0.803)};
\addlegendentry{SLA-first}
\addplot[thick, color=teal, mark=none] coordinates {(2.722,0.814) (3.033,0.837) (3.475,0.860)};
\end{axis}
\end{tikzpicture}

%% file: paper/figures/pareto_frontier_fiqa.tex
\begin{tikzpicture}
\begin{axis}[
width=0.95\linewidth,
height=0.55\linewidth,
xlabel={p95 latency (s)},
ylabel={Readiness score},
ymin=0,
ymax=1,
grid=major,
legend style={font=\small, at={(0.5,-0.22)}, anchor=north, legend columns=3},
tick label style={font=\small},
label style={font=\small},
]
\addplot[only marks, mark=o, mark size=2.2pt, color=blue] coordinates {(4.504,0.508) (4.061,0.492) (3.534,0.543) (3.327,0.711) (3.582,0.674) (3.552,0.696) (4.193,0.646) (3.687,0.625) (3.206,0.660) (5.197,0.533) (4.597,0.517) (4.361,0.552) (3.790,0.744) (5.318,0.688) (3.398,0.718) (4.300,0.702) (4.629,0.648) (4.196,0.685) (9.817,0.449) (10.811,0.424) (10.642,0.425) (8.795,0.652) (9.989,0.604) (10.344,0.583) (9.214,0.588) (9.522,0.549) (8.878,0.547)};
\addlegendentry{Cost-first}
\addplot[thick, color=blue, mark=none] coordinates {(3.206,0.660) (3.327,0.711) (3.398,0.718) (3.790,0.744)};
\addplot[only marks, mark=square*, mark size=2.2pt, color=orange] coordinates {(4.062,0.674) (4.403,0.676) (4.228,0.691) (4.164,0.759) (3.894,0.723) (3.908,0.739) (3.687,0.749) (3.680,0.709) (3.625,0.732) (4.600,0.642) (3.958,0.624) (3.329,0.668) (4.155,0.735) (4.049,0.699) (3.654,0.708) (3.681,0.703) (3.577,0.676) (3.169,0.722) (9.211,0.567) (9.411,0.528) (9.522,0.538) (8.210,0.659) (9.223,0.596) (9.890,0.579) (9.058,0.621) (9.399,0.589) (10.133,0.589)};
\addlegendentry{Risk-first}
\addplot[thick, color=orange, mark=none] coordinates {(3.169,0.722) (3.625,0.732) (3.687,0.749) (4.164,0.759)};
\addplot[only marks, mark=triangle*, mark size=2.2pt, color=teal] coordinates {(4.007,0.726) (3.782,0.710) (3.577,0.740) (3.325,0.814) (3.476,0.776) (3.254,0.783) (4.094,0.789) (3.273,0.763) (3.377,0.793) (4.909,0.744) (4.504,0.743) (4.704,0.761) (3.890,0.829) (4.566,0.807) (4.264,0.806) (4.075,0.818) (4.048,0.783) (3.977,0.824) (8.702,0.684) (9.517,0.652) (8.800,0.653) (9.165,0.751) (7.818,0.715) (10.749,0.716) (9.010,0.724) (8.697,0.697) (9.415,0.708)};
\addlegendentry{SLA-first}
\addplot[thick, color=teal, mark=none] coordinates {(3.254,0.783) (3.325,0.814) (3.890,0.829)};
\end{axis}
\end{tikzpicture}

%% file: paper/tables/retriever_ablation_table.tex
\begin{table}[!ht]
\centering
\caption{Retriever ablation (SciFact; sla-first; k=5; seed=2024).}
\label{tab:retriever-ablation}
\scriptsize
\setlength{\tabcolsep}{3pt}
\resizebox{\linewidth}{!}{%
\begin{tabular}{l r r l l r r r}
\toprule
Scenario & k & Seed & Retriever & Reranker & Hit@k & Faithfulness & p95 latency (ms) \\
\midrule
SLA-first & 5 & 2024 & bm25 & off & 0.850 & 0.479 & 3054 \\
SLA-first & 5 & 2024 & bm25 & on & 0.812 & 0.525 & 2859 \\
SLA-first & 5 & 2024 & dense & off & 0.812 & 0.470 & 3392 \\
SLA-first & 5 & 2024 & dense & on & 0.800 & 0.481 & 3229 \\
\bottomrule
\end{tabular}
}
\end{table}

%% file: paper/tables/seed_stats_table.tex
\begin{table}[!ht]
\centering
\caption{Seeded robustness (mean $\pm$ std; p95 in ms; SciFact and FiQA).}
\label{tab:seed-stats}
\scriptsize
\setlength{\tabcolsep}{2pt}
\resizebox{\linewidth}{!}{%
\begin{tabular}{l l l r r r r r}
\toprule
Dataset & Model & Scenario & k & Runs & Faith. & Hit@k & p95 \\
\midrule
FiQA & gpt-4.1 & Cost-first & 3 & 3 & 0.65 $\pm$ 0.04 & 0.53 $\pm$ 0.08 & 3487 $\pm$ 139 \\
FiQA & gpt-4.1 & Cost-first & 5 & 3 & 0.65 $\pm$ 0.06 & 0.57 $\pm$ 0.06 & 3695 $\pm$ 494 \\
FiQA & gpt-4.1 & Risk-first & 3 & 3 & 0.65 $\pm$ 0.01 & 0.53 $\pm$ 0.08 & 3953 $\pm$ 264 \\
FiQA & gpt-4.1 & Risk-first & 5 & 3 & 0.67 $\pm$ 0.06 & 0.57 $\pm$ 0.06 & 3476 $\pm$ 271 \\
FiQA & gpt-4.1 & SLA-first & 3 & 3 & 0.61 $\pm$ 0.04 & 0.53 $\pm$ 0.08 & 3352 $\pm$ 113 \\
FiQA & gpt-4.1 & SLA-first & 5 & 3 & 0.67 $\pm$ 0.05 & 0.57 $\pm$ 0.06 & 3581 $\pm$ 447 \\
FiQA & gpt-4.1-mini & Cost-first & 3 & 3 & 0.73 $\pm$ 0.06 & 0.53 $\pm$ 0.08 & 4169 $\pm$ 1014 \\
FiQA & gpt-4.1-mini & Cost-first & 5 & 3 & 0.79 $\pm$ 0.05 & 0.57 $\pm$ 0.06 & 4375 $\pm$ 226 \\
FiQA & gpt-4.1-mini & Risk-first & 3 & 3 & 0.72 $\pm$ 0.03 & 0.53 $\pm$ 0.08 & 3989 $\pm$ 152 \\
FiQA & gpt-4.1-mini & Risk-first & 5 & 3 & 0.75 $\pm$ 0.02 & 0.57 $\pm$ 0.06 & 3664 $\pm$ 34 \\
FiQA & gpt-4.1-mini & SLA-first & 3 & 3 & 0.71 $\pm$ 0.01 & 0.53 $\pm$ 0.08 & 4240 $\pm$ 339 \\
FiQA & gpt-4.1-mini & SLA-first & 5 & 3 & 0.78 $\pm$ 0.07 & 0.57 $\pm$ 0.06 & 4033 $\pm$ 51 \\
FiQA & gpt-5.2 & Cost-first & 3 & 3 & 0.39 $\pm$ 0.07 & 0.53 $\pm$ 0.08 & 9709 $\pm$ 811 \\
FiQA & gpt-5.2 & Cost-first & 5 & 3 & 0.39 $\pm$ 0.05 & 0.57 $\pm$ 0.06 & 9205 $\pm$ 322 \\
FiQA & gpt-5.2 & Risk-first & 3 & 3 & 0.36 $\pm$ 0.07 & 0.53 $\pm$ 0.08 & 9108 $\pm$ 846 \\
FiQA & gpt-5.2 & Risk-first & 5 & 3 & 0.39 $\pm$ 0.02 & 0.57 $\pm$ 0.06 & 9530 $\pm$ 549 \\
FiQA & gpt-5.2 & SLA-first & 3 & 3 & 0.37 $\pm$ 0.04 & 0.53 $\pm$ 0.08 & 9244 $\pm$ 1467 \\
FiQA & gpt-5.2 & SLA-first & 5 & 3 & 0.39 $\pm$ 0.02 & 0.57 $\pm$ 0.06 & 9041 $\pm$ 360 \\
SciFact & gpt-4.1 & Cost-first & 3 & 3 & 0.58 $\pm$ 0.09 & 0.82 $\pm$ 0.03 & 2933 $\pm$ 148 \\
SciFact & gpt-4.1 & Cost-first & 5 & 3 & 0.58 $\pm$ 0.14 & 0.84 $\pm$ 0.01 & 3585 $\pm$ 408 \\
SciFact & gpt-4.1 & Risk-first & 3 & 3 & 0.56 $\pm$ 0.07 & 0.82 $\pm$ 0.03 & 3304 $\pm$ 278 \\
SciFact & gpt-4.1 & Risk-first & 5 & 3 & 0.59 $\pm$ 0.10 & 0.84 $\pm$ 0.01 & 3049 $\pm$ 224 \\
SciFact & gpt-4.1 & SLA-first & 3 & 3 & 0.55 $\pm$ 0.07 & 0.82 $\pm$ 0.03 & 3154 $\pm$ 504 \\
SciFact & gpt-4.1 & SLA-first & 5 & 3 & 0.58 $\pm$ 0.15 & 0.84 $\pm$ 0.01 & 3200 $\pm$ 244 \\
SciFact & gpt-4.1-mini & Cost-first & 3 & 3 & 0.58 $\pm$ 0.15 & 0.82 $\pm$ 0.03 & 3516 $\pm$ 360 \\
SciFact & gpt-4.1-mini & Cost-first & 5 & 3 & 0.63 $\pm$ 0.10 & 0.84 $\pm$ 0.01 & 3324 $\pm$ 143 \\
SciFact & gpt-4.1-mini & Risk-first & 3 & 3 & 0.58 $\pm$ 0.10 & 0.82 $\pm$ 0.03 & 3744 $\pm$ 338 \\
SciFact & gpt-4.1-mini & Risk-first & 5 & 3 & 0.67 $\pm$ 0.04 & 0.84 $\pm$ 0.01 & 3736 $\pm$ 342 \\
SciFact & gpt-4.1-mini & SLA-first & 3 & 3 & 0.58 $\pm$ 0.14 & 0.82 $\pm$ 0.03 & 3439 $\pm$ 79 \\
SciFact & gpt-4.1-mini & SLA-first & 5 & 3 & 0.63 $\pm$ 0.08 & 0.84 $\pm$ 0.01 & 3505 $\pm$ 159 \\
SciFact & gpt-5.2 & Cost-first & 3 & 3 & 0.62 $\pm$ 0.07 & 0.82 $\pm$ 0.03 & 6224 $\pm$ 217 \\
SciFact & gpt-5.2 & Cost-first & 5 & 3 & 0.58 $\pm$ 0.09 & 0.84 $\pm$ 0.01 & 6474 $\pm$ 626 \\
SciFact & gpt-5.2 & Risk-first & 3 & 3 & 0.60 $\pm$ 0.09 & 0.82 $\pm$ 0.03 & 7072 $\pm$ 987 \\
SciFact & gpt-5.2 & Risk-first & 5 & 3 & 0.56 $\pm$ 0.11 & 0.84 $\pm$ 0.01 & 6547 $\pm$ 80 \\
SciFact & gpt-5.2 & SLA-first & 3 & 3 & 0.61 $\pm$ 0.07 & 0.82 $\pm$ 0.03 & 6792 $\pm$ 1152 \\
SciFact & gpt-5.2 & SLA-first & 5 & 3 & 0.58 $\pm$ 0.05 & 0.84 $\pm$ 0.01 & 6229 $\pm$ 579 \\
\bottomrule
\end{tabular}
}
\end{table}

%% file: paper/figures/seed_robustness.tex
\begin{tikzpicture}
\begin{axis}[
width=0.95\linewidth,
height=0.45\linewidth,
xlabel={Scenario},
ylabel={Readiness score (mean $\pm$ std)},
symbolic x coords={cost-first,risk-first,sla-first},
xtick=data,
xticklabels={Cost-first,Risk-first,SLA-first},
ymin=0,
ymax=1,
grid=major,
legend style={font=\small, at={(0.5,-0.26)}, anchor=north, legend columns=3},
tick label style={font=\small},
label style={font=\small},
]
\addplot+[mark=o, color=orange, mark size=2.2pt, error bars/.cd, y dir=both, y explicit] coordinates {(cost-first,0.644)+-(0,0.018) (risk-first,0.700)+-(0,0.023) (sla-first,0.782)+-(0,0.016)};
\addlegendentry{FiQA | gpt-4.1}
\addplot+[mark=square*, color=orange, mark size=2.2pt, error bars/.cd, y dir=both, y explicit] coordinates {(cost-first,0.678)+-(0,0.028) (risk-first,0.730)+-(0,0.020) (sla-first,0.808)+-(0,0.022)};
\addlegendentry{FiQA | gpt-4.1-mini}
\addplot+[mark=triangle*, color=orange, mark size=2.2pt, error bars/.cd, y dir=both, y explicit] coordinates {(cost-first,0.561)+-(0,0.023) (risk-first,0.600)+-(0,0.018) (sla-first,0.710)+-(0,0.014)};
\addlegendentry{FiQA | gpt-5.2}
\addplot+[mark=o, color=blue, mark size=2.2pt, error bars/.cd, y dir=both, y explicit] coordinates {(cost-first,0.682)+-(0,0.031) (risk-first,0.737)+-(0,0.030) (sla-first,0.798)+-(0,0.033)};
\addlegendentry{SciFact | gpt-4.1}
\addplot+[mark=square*, color=blue, mark size=2.2pt, error bars/.cd, y dir=both, y explicit] coordinates {(cost-first,0.695)+-(0,0.021) (risk-first,0.764)+-(0,0.010) (sla-first,0.811)+-(0,0.017)};
\addlegendentry{SciFact | gpt-4.1-mini}
\addplot+[mark=triangle*, color=blue, mark size=2.2pt, error bars/.cd, y dir=both, y explicit] coordinates {(cost-first,0.677)+-(0,0.019) (risk-first,0.724)+-(0,0.035) (sla-first,0.797)+-(0,0.010)};
\addlegendentry{SciFact | gpt-5.2}
\end{axis}
\end{tikzpicture}

%% file: paper/sections/06-validation.tex
Validation focuses on automated sanity checks and regression detection over
artifact-complete runs. For ticket workflows, we run seeded regression variants
on Azure (\texttt{sample\_n=60} per dataset): a \emph{bias} variant that shifts
routing behavior, and a \emph{policy} variant that removes key safety constraints.
Each candidate is compared against the latest baseline using CI thresholds.

Across both ticket datasets (CS and IT), the policy regression reliably triggers
hard-gate failures with large drops in policy pass and workflow success
($\Delta$Policy = -0.97 on CS and -0.93 on IT in the latest suite).
In the current Azure regression set, the bias variant also fails gates in both
datasets (smaller policy/workflow deltas than policy-variant failures, but still
beyond configured thresholds): CS bias shifts routing (+0.07) with
$\Delta$Policy=-0.08, and IT bias produces smaller quality shifts
($\Delta$Routing=-0.01, $\Delta$Policy=-0.02) but remains gate-failing.
Table~\ref{tab:ticket-regressions} summarizes deltas and gate outcomes;
per-error-type breakdowns are provided in
Tables~\ref{tab:ticket-error-summary} and \ref{tab:ticket-error-full}
(Appendix~\ref{app:ticket-error-full}).

\IfFileExists{paper/tables/ticket_regression_gate.tex}{\input{paper/tables/ticket_regression_gate.tex}}{}

We do not include a human validation study in this work. Given the updated
hosted-model scale, the next validation increment is a targeted manual audit on
high-disagreement slices (model disagreement, gate-borderline cases, and long-tail
latency outliers).

%% file: paper/tables/ticket_regression_gate.tex
\begin{table}[!ht]
\centering
\caption{Seeded prompt regressions and CI gate outcome (T1/T2).}
\label{tab:ticket-regressions}
\scriptsize
\setlength{\tabcolsep}{3pt}
\begin{tabular}{l l r r r r l}
\toprule
Dataset & Variant & $\Delta$Workflow & $\Delta$Policy & $\Delta$Routing & $\Delta$p95 (ms) & Gate \\
\midrule
CS & bias & 0.03 & -0.08 & 0.07 & -120 & fail \\
CS & policy & -0.27 & -0.97 & 0.07 & -371 & fail \\
IT & bias & -0.03 & -0.02 & -0.01 & -293 & fail \\
IT & policy & -0.33 & -0.93 & 0.02 & 13 & fail \\
\bottomrule
\end{tabular}
\end{table}

%% file: paper/sections/07-lessons.tex
Four practical lessons emerged from the Azure matrix.
First, newer/larger model classes are not automatically superior under a fixed
prompt+schema contract: in our BEIR setup, \texttt{gpt-4.1-mini} is strongest on
FiQA readiness/faithfulness (0.777 $\pm$ 0.069), while \texttt{gpt-5.2} shows
a 50\% drop in faithfulness (0.387 $\pm$ 0.016) and carries a clear latency
penalty. We hypothesize that \texttt{gpt-5.2} may synthesize and paraphrase
retrieved passages more aggressively rather than grounding answers close to the
source text---behavior that RAGAS faithfulness penalizes---and that the fixed
prompt contract was not tuned for the newer model's instruction-following style.
This underscores the need to re-evaluate prompts when upgrading model versions,
even within the same provider family.
Second, retrieval depth is a true operating knob: higher k consistently raises
hit@k, but cost/latency growth can erase readiness gains under cost/SLA-weighted
gates. Third, prompt updates are dataset-dependent: ticket prompt v2 improves
latency on both datasets but improves routing mainly on IT (not CS), so
"single winner" prompt assumptions are brittle. Fourth, regression gates remain
high-leverage: policy-hard constraints
catch harmful prompt variants that would pass if we monitored only aggregate
quality metrics.

Operationally, worker parallelism plus checkpoint/resume support was essential:
it allowed recovery from machine restarts and connectivity failures without
discarding partial runs, which materially reduced experiment turnaround time.

%% file: paper/sections/08-limitations.tex
Our evaluation relies on public datasets and reference-free metrics that do not
fully capture production behavior. BEIR is predominantly English and emphasizes
retrieval-grounding proxies rather than end-task business outcomes. Cost and
latency here are measured at the harness boundary and should be validated in
production telemetry with end-user traffic patterns. A small number of runs
exhibited extreme p95 latency values (e.g., 32--33s for \texttt{gpt-4.1}
in isolated seeds), consistent with transient Azure connectivity timeouts
rather than intrinsic model behavior; such outliers can inflate p95 estimates
and should be investigated via OTel trace inspection in production deployments.

Although the core Azure matrix is complete (multi-seed, multi-scenario,
multi-model), external validity remains bounded: runs were executed in one cloud
provider/region configuration and with fixed prompt contracts. Ticket experiments
were concentrated on \texttt{gpt-4.1}; cross-model ticket calibration is still
pending.

This paper does not provide a controlled cross-provider comparison. Results are
reported for Azure-hosted OpenAI deployments under fixed prompt and retrieval
contracts; cross-provider transferability should be tested in future work before
generalization.

Trace artifacts also require strict redaction and retention controls before
production deployment.

For synthetic data generation, LLM-based augmentation can introduce memorization
risks and contamination of evaluation data \citep{carlini2020extracting,contamination-quiz-2023,livecodebench-2024},
and may bias outputs toward the generator's style or amplify label noise.
Model collapse concerns apply when synthetic data is used for training
\citep{model-collapse-2024}. Our pipeline mitigates these with neutral prompts,
schema and policy checks, deduplication, similarity filtering, and controlled
distributions, but residual risks remain; synthetic data should be treated as a
complement to (not a replacement for) real-world validation.

%% file: paper/sections/09-related.tex
\begingroup\sloppy
We build on reference-free RAG evaluation (Ragas) \citep{ragas2023}, vendor-neutral
observability standards (OpenTelemetry) \citep{opentelemetry-docs}, CI evaluation tooling
(promptfoo) \citep{promptfoo-ci,promptfoo-http}, and the BEIR benchmark for retrieval
evaluation \citep{beir2021}. We differentiate by combining these components into a
single readiness workflow with explicit cost and latency instrumentation, run
artifacts, and decision-oriented frontiers. This section positions our harness
relative to existing eval frameworks and operational evaluation practices.
Many RAG evaluation toolkits already exist, but they typically optimize or report
quality signals in isolation rather than enforcing deployment readiness with
observability and CI gating.

Holistic evaluation suites such as HELM \citep{helm2023} emphasize multi-metric,
scenario-level benchmarking for language models, highlighting the need to surface
trade-offs beyond accuracy. ARES introduces automated component-level evaluation
for RAG systems using lightweight LM judges \citep{ares2024}. We align with these
efforts on multi-signal evaluation, but focus on release readiness by binding
evaluation to instrumentation, artifacts, and CI gates. Experience reports on RAG
engineering failure points \citep{rag-failure-points} motivate our emphasis on
observability and policy/grounding checks as first-class readiness signals.

Recent work on testing LLM applications proposes a layered testing perspective and
a lightweight interaction protocol (AICL) for replayable evaluation
\citep{rethinking-testing-llm}. We align with its emphasis on system-level testing
while focusing on deployment readiness gates and observability artifacts. Work on
output drift in financial workflows highlights the need for deterministic controls
and traceable artifacts in regulated settings \citep{llm-output-drift}; our harness
generalizes these ideas beyond finance and couples them with scenario-weighted
trade-off analysis. Economic evaluation of LLMs formalizes accuracy--cost trade-offs
using economic constraints \citep{economic-eval-llm}; we adopt Pareto reporting for
decision support while keeping weights explicit and auditable. BEIR infrastructure
updates and leaderboards further motivate reproducible retrieval baselines
\citep{beir-resources}. A later companion study on automated self-testing quality
gates \citep{maiorano2026selftesting} narrows the scope from cross-scenario readiness
assessment to longitudinal release governance for a single internally deployed
multi-agent application. We view that work as a specialization of the broader
evaluation-and-observability layer introduced here.
\endgroup

%% file: paper/sections/10-conclusion.tex
We present an evaluation and observability harness that turns LLM/RAG readiness
into measurable, auditable signals with CI gates and Pareto-style deployment
trade-offs. With the full Azure matrix now executed for BEIR (SciFact/FiQA),
the harness supports model-by-model decisions under cost-first, risk-first, and
SLA-first scoring without mixing runs across providers or model families.

The current evidence shows that model choice, retrieval depth, and policy gates
interact in non-trivial ways: FiQA favors \texttt{gpt-4.1-mini} for
faithfulness/readiness, while SciFact is more tightly clustered and
\texttt{gpt-5.2} mainly pays a latency premium. Ticket CI gates also proved
effective, rejecting all latest bias/policy regression candidates.
Reproducible artifacts are therefore necessary to avoid over-generalized
conclusions from single metrics. Future work includes broader multilingual and
industry datasets, stronger adversarial validation, and human-audit layers on
disagreement slices, plus cross-provider replication under matched protocols.
Another immediate direction is to specialize this general readiness layer into
task-specific release workflows, such as automated self-testing quality gates for
agentic applications \citep{maiorano2026selftesting}, while keeping shared
observability and artifact contracts stable across systems.

%% file: paper/tables/summary_table.tex
\begin{table}[H]
\centering
\caption{Top runs by readiness score (compact view; dataset: SciFact).}
\label{tab:summary}
\scriptsize
\setlength{\tabcolsep}{3pt}
\resizebox{\linewidth}{!}{%
\begin{tabular}{l l r r r r r r}
\toprule
Scenario & Model & k & Seed & Hit@k & Faithfulness & Score & p95 latency (ms) \\
\midrule
SLA-first & gpt-4.1-mini & 3 & 42 & 0.817 & 0.727 & 0.860 & 3475 \\
SLA-first & gpt-5.2 & 3 & 42 & 0.817 & 0.686 & 0.847 & 6455 \\
SLA-first & gpt-4.1 & 3 & 42 & 0.817 & 0.628 & 0.837 & 3033 \\
SLA-first & gpt-4.1 & 5 & 42 & 0.833 & 0.752 & 0.837 & 3313 \\
SLA-first & gpt-5.2 & 3 & 2024 & 0.850 & 0.597 & 0.833 & 8075 \\
SLA-first & gpt-4.1-mini & 5 & 42 & 0.833 & 0.724 & 0.830 & 3379 \\
SLA-first & gpt-4.1-mini & 3 & 1337 & 0.783 & 0.559 & 0.818 & 3348 \\
SLA-first & gpt-4.1 & 3 & 2024 & 0.850 & 0.501 & 0.814 & 2722 \\
SLA-first & gpt-5.2 & 3 & 1337 & 0.783 & 0.552 & 0.812 & 5847 \\
SLA-first & gpt-4.1 & 3 & 1337 & 0.783 & 0.527 & 0.810 & 3708 \\
\bottomrule
\end{tabular}
}
\end{table}

%% file: paper/tables/fiqa_summary_table.tex
\begin{table}[H]
\centering
\caption{Top runs by readiness score (compact view; dataset: FiQA).}
\label{tab:fiqa-summary}
\scriptsize
\setlength{\tabcolsep}{3pt}
\resizebox{\linewidth}{!}{%
\begin{tabular}{l l r r r r r r}
\toprule
Scenario & Model & k & Seed & Hit@k & Faithfulness & Score & p95 latency (ms) \\
\midrule
SLA-first & gpt-4.1-mini & 3 & 1337 & 0.617 & 0.713 & 0.829 & 3890 \\
SLA-first & gpt-4.1-mini & 5 & 42 & 0.567 & 0.848 & 0.824 & 3977 \\
SLA-first & gpt-4.1-mini & 5 & 1337 & 0.633 & 0.775 & 0.818 & 4075 \\
SLA-first & gpt-4.1 & 3 & 1337 & 0.617 & 0.647 & 0.814 & 3325 \\
SLA-first & gpt-4.1-mini & 3 & 2024 & 0.483 & 0.710 & 0.807 & 4566 \\
SLA-first & gpt-4.1-mini & 3 & 42 & 0.483 & 0.698 & 0.806 & 4264 \\
SLA-first & gpt-4.1 & 5 & 42 & 0.567 & 0.717 & 0.793 & 3377 \\
SLA-first & gpt-4.1 & 5 & 1337 & 0.633 & 0.654 & 0.789 & 4094 \\
SLA-first & gpt-4.1-mini & 5 & 2024 & 0.517 & 0.709 & 0.783 & 4048 \\
SLA-first & gpt-4.1 & 3 & 42 & 0.483 & 0.600 & 0.783 & 3254 \\
\bottomrule
\end{tabular}
}
\end{table}

%% file: paper/tables/weight_ablation_table.tex
\begin{table}[H]
\centering
\caption{Readiness score weight ablation (top-1 agreement and top-5 overlap).}
\label{tab:weight-ablation}
\begin{tabular}{l l c c}
\toprule
Dataset & Scenario & Top-1 (U/NC/PG) & Top-5 overlap (U/NC/PG) \\
\midrule
FiQA & Cost-first & Y/N/Y & 4/5 | 1/5 | 5/5 \\
FiQA & Risk-first & Y/N/Y & 4/5 | 1/5 | 5/5 \\
FiQA & SLA-first & Y/N/Y & 5/5 | 2/5 | 5/5 \\
SciFact & Cost-first & Y/N/Y & 5/5 | 1/5 | 5/5 \\
SciFact & Risk-first & Y/N/Y & 3/5 | 4/5 | 5/5 \\
SciFact & SLA-first & Y/N/Y & 5/5 | 2/5 | 5/5 \\
CS Tickets & Cost-first & Y/Y/Y & 1/1 | 1/1 | 1/1 \\
CS Tickets & Risk-first & Y/Y/N & 5/5 | 5/5 | 2/5 \\
IT Tickets & Cost-first & Y/Y/Y & 1/1 | 1/1 | 1/1 \\
IT Tickets & Risk-first & Y/Y/N & 4/5 | 5/5 | 3/5 \\
\bottomrule
\end{tabular}
\end{table}

%% file: paper/tables/ticket_summary.tex
\begin{table}[H]
\centering
\caption{Ticket routing results (T1/T2; compact view).}
\label{tab:ticket-results}
\small
\setlength{\tabcolsep}{2pt}
\begin{tabular}{>{\raggedright\arraybackslash}p{4.2cm} r r r r r r}
\toprule
Run & N & R Acc & Macro F1 & Policy & Workflow & p95 ms \\
\midrule
Prompt v1 (seed 1337) (CS) & 200 & 0.325 & 0.280 & 0.985 & 0.215 & 3203 \\
Prompt v1 (seed 1337) (IT) & 200 & 0.410 & 0.359 & 0.965 & 0.390 & 2912 \\
Prompt v2 (seed 1337) (CS) & 200 & 0.305 & 0.269 & 0.890 & 0.165 & 2948 \\
Prompt v2 (seed 42) (IT) & 200 & 0.430 & 0.369 & 0.905 & 0.375 & 2350 \\
\bottomrule
\end{tabular}
\end{table}

%% file: paper/tables/ticket_seed_stats.tex
\begin{table}[H]
\centering
\caption{Ticket seed robustness (mean $\pm$ std across seeded prompt runs; provider/model explicit).}
\label{tab:ticket-seed-stats}
\resizebox{\linewidth}{!}{%
\begin{tabular}{l l l l r r r r r}
\hline
Dataset & Prompt & Provider & Model & Runs & Routing Acc & Macro F1 & Workflow & p95 latency (ms) \\
\hline
Customer support (CS) & v1 & azure & gpt-4.1 & 3 & 0.295 $\pm$ 0.033 & 0.204 $\pm$ 0.068 & 0.167 $\pm$ 0.053 & 3444 $\pm$ 292 \\
Customer support (CS) & v2 & azure & gpt-4.1 & 3 & 0.292 $\pm$ 0.013 & 0.215 $\pm$ 0.050 & 0.148 $\pm$ 0.015 & 3150 $\pm$ 194 \\
IT tickets (IT) & v1 & azure & gpt-4.1 & 3 & 0.393 $\pm$ 0.029 & 0.351 $\pm$ 0.010 & 0.362 $\pm$ 0.037 & 2743 $\pm$ 146 \\
IT tickets (IT) & v2 & azure & gpt-4.1 & 3 & 0.408 $\pm$ 0.026 & 0.377 $\pm$ 0.009 & 0.358 $\pm$ 0.038 & 2447 $\pm$ 93 \\
\hline
\end{tabular}
}
\end{table}

%% file: paper/sections/appendix-extras.tex
This appendix provides concrete artifacts for reproducibility: exemplar prompts,
schemas, pseudocode, and trace excerpts used by the readiness harness.

\subsection{Ticket routing prompt (T1/T2)}
\begin{codebox}
System:
You are a strict routing classifier for customer support tickets.
Return only valid JSON that matches the schema.

User:
Ticket: "My login fails with error 401 after resetting my password."
Labels: ["account_support", "technical_support", "billing", "product_support"]
Policy: Do not request passwords, secrets, or PII.
\end{codebox}

\noindent\textbf{Prompt v1 $\rightarrow$ v2 changes.}
The ticket-routing study used two prompt variants. Prompt v1 lists label definitions inline as English
descriptions. Prompt v2 replaces those inline definitions with a decision-guidance block (escalation
heuristics and policy reminders) and adds an explicit JSON schema with \texttt{additionalProperties: false}.
The reduced public reproducibility artifact does not ship the full internal prompt files; instead, this
appendix provides a representative excerpt and the released tables report the measured effect of the
prompt revision. The tighter output constraint in v2 reduces
LLM token generation and lowers p95 latency on both datasets (+11\% IT, +9\% CS), while improving
routing accuracy on IT (+1.5 pp) and slightly reducing it on CS ($-$0.3 pp)—consistent with the
finding that routing gains are dataset-dependent.

\subsection{Routing output schema (JSON)}
\begin{codebox}
{
  "type": "object",
  "properties": {
    "route_label": {"type": "string"},
    "confidence": {"type": "number", "minimum": 0, "maximum": 1},
    "should_escalate": {"type": "boolean"},
    "policy_violations": {"type": "array", "items": {"type": "string"}}
  },
  "required": ["route_label", "confidence", "should_escalate", "policy_violations"]
}
\end{codebox}

\subsection{Batch runner pseudocode}
\begin{codebox}
for scenario in ["risk-first", "cost-first", "sla-first"]:
  for top_k in [3, 5, 10]:
    sample = dataset.sample(n=60, seed=seed)
    results = []
    for item in sample:
      response = call_infer_api(item, top_k, scenario)
      record_metrics(response)
      results.append(response)
    # resume/checkpoint: load prior rows before rerun
    # worker pool + delay controls API pressure
    write_run_artifacts(results)
    run_ragas(results)
    update_summary_tables()
\end{codebox}

\subsection{Example trace excerpt}
\begin{codebox}
infer (run_id=azure_core_beir_scifact_gpt-4.1_sla-first_k5_seed42_..., scenario=sla-first, latency_ms=4285, cost_usd_est=0.0114)
  route.classify (predicted_label=technical_support, latency_ms=412)
  rag.retrieve (top_k=5, index_id=beir_scifact, latency_ms=58)
  rag.generate (model_id=gpt-4.1, provider=azure, cache_hit=false, latency_ms=3611)
  validate.policy (pass=true, latency_ms=12)
  respond.finalize (should_escalate=false, latency_ms=9)
\end{codebox}

\subsection{Regression example}
\begin{codebox}
Change: prompt update removed a hard policy reminder.
Observed: policy_violations rose from 0.00 to 0.12 in CI gate.
Action: revert prompt change and add promptfoo gate to prevent regressions.
\end{codebox}

\subsection{Dataset examples}
\label{app:dataset-examples}
\begin{codebox}
Ticket (IT Service, synthetic):
Text: "VPN drops after 5 minutes; reconnect fails on Windows 11."
Label: "access"
\end{codebox}

\begin{codebox}
Ticket (Multilingual, synthetic):
Text (de): "Login funktioniert nicht, Fehler 403 nach Passwort-Reset."
Label: "account_support"
\end{codebox}

\begin{codebox}
SciFact (BEIR, synthetic):
Query: "Does vitamin D reduce the risk of influenza?"
Evidence snippet: "We found no significant reduction in influenza incidence..."
\end{codebox}

\begin{codebox}
FiQA (BEIR, synthetic):
Query: "What is the 50/30/20 budgeting rule?"
Passage snippet: "A common guideline allocates 50
\end{codebox}

\subsection{Synthetic generation example}
\begin{codebox}
Synthetic ticket (LLM, es):
{
  "ticket_id": "synllm_42_00001",
  "language": "es",
  "channel": "chat",
  "priority": "high",
  "queue": "technical_support",
  "summary": "la conexion VPN se corta durante sincronizacion (chat)",
  "description": "El usuario reporta que la conexion VPN se corta durante la sincronizacion de datos. El problema persiste tras reiniciar. Impacto: retrasa reportes. Canal: chat. Prioridad: alto. Producto: integrations.",
  "requester_type": "user",
  "product_area": "integrations",
  "policy_flags": [],
  "should_escalate": true,
  "escalation_reason": "data_loss"
}
\end{codebox}

\begin{codebox}
Quality snapshot (LLM, N=500, strict):
schema_error_rate=0.00
policy_violation_rate=0.00
lexical_diversity=0.0898
unique_summary_rate=0.3960
description_len_mean=212.8
\end{codebox}

\subsection{Artifact format examples}
\label{app:artifact-examples}
\begin{codebox}
summary.csv (one row)
run_id,dataset_id,scenario,top_k,score,faithfulness,answer_relevance,p95_latency_ms
azure_core_beir_scifact_gpt-4.1-mini_sla-first_k5_seed42_20260220_085315,beir_scifact,sla-first,5,0.830,0.724,,3379
\end{codebox}

\begin{codebox}
report.json (excerpt)
{
  "run_id": "azure_core_beir_scifact_gpt-4.1-mini_sla-first_k5_seed42_20260220_085315",
  "dataset_id": "beir_scifact",
  "scenario": "sla-first",
  "llm": {"provider": "azure", "model": "gpt-4.1-mini"},
  "metrics": {"faithfulness": 0.724, "answer_relevance": null},
  "latency_ms": {"p95": 3379}
}
\end{codebox}

%% file: paper/tables/ticket_summary_full.tex
\newcommand{\runA}{TF-IDF baseline (CS)}
\newcommand{\runB}{Prompt v1 (seed 1337) (CS)}
\newcommand{\runC}{Prompt v1 (seed 2024) (CS)}
\newcommand{\runD}{Prompt v1 (seed 42) (CS)}
\newcommand{\runE}{Prompt v2 (seed 1337) (CS)}
\newcommand{\runF}{Prompt v2 (seed 2024) (CS)}
\newcommand{\runG}{Prompt v2 (seed 42) (CS)}
\newcommand{\runH}{Regression:\allowbreak{} baseline (CS)}
\newcommand{\runI}{Regression:\allowbreak{} bias (CS)}
\newcommand{\runJ}{Regression:\allowbreak{} policy (CS)}
\newcommand{\runK}{TF-IDF baseline (IT)}
\newcommand{\runL}{Prompt v1 (seed 1337) (IT)}
\newcommand{\runM}{Prompt v1 (seed 2024) (IT)}
\newcommand{\runN}{Prompt v1 (seed 42) (IT)}
\newcommand{\runO}{Prompt v2 (seed 1337) (IT)}
\newcommand{\runP}{Prompt v2 (seed 2024) (IT)}
\newcommand{\runQ}{Prompt v2 (seed 42) (IT)}
\newcommand{\runR}{Regression:\allowbreak{} baseline (IT)}
\newcommand{\runS}{Regression:\allowbreak{} bias (IT)}
\newcommand{\runT}{Regression:\allowbreak{} policy (IT)}
\newcommand{\modelA}{TF-IDF}
\newcommand{\modelB}{gpt-4.1 (azure)}
\begin{table}[H]
\centering
\caption{Ticket routing results (T1/T2; full).}
\label{tab:ticket-results-full}
\small
\setlength{\tabcolsep}{0.5pt}
\resizebox{\linewidth}{!}{%
\begin{tabular}{@{}l l r r r r r r@{}}
\toprule
Run & Model & N & R Acc & F1 & Policy & Workflow & p95 ms \\
\midrule
\runA & \modelA & 200 & 0.500 & 0.481 & - & 0.500 & 0 \\
\runB & \modelB & 200 & 0.325 & 0.280 & 0.985 & 0.215 & 3203 \\
\runC & \modelB & 200 & 0.300 & 0.183 & 0.965 & 0.175 & 3769 \\
\runD & \modelB & 200 & 0.260 & 0.150 & 0.970 & 0.110 & 3360 \\
\runE & \modelB & 200 & 0.305 & 0.269 & 0.890 & 0.165 & 2948 \\
\runF & \modelB & 200 & 0.280 & 0.203 & 0.890 & 0.145 & 3169 \\
\runG & \modelB & 200 & 0.290 & 0.172 & 0.890 & 0.135 & 3334 \\
\runH & \modelB & 60 & 0.283 & 0.212 & 0.967 & 0.267 & 3289 \\
\runI & \modelB & 60 & 0.350 & 0.161 & 0.883 & 0.300 & 3169 \\
\runJ & \modelB & 60 & 0.350 & 0.239 & 0.000 & 0.000 & 2918 \\
\runK & \modelA & 200 & 0.880 & 0.889 & - & 0.880 & 0 \\
\runL & \modelB & 200 & 0.410 & 0.359 & 0.965 & 0.390 & 2912 \\
\runM & \modelB & 200 & 0.360 & 0.340 & 0.940 & 0.320 & 2661 \\
\runN & \modelB & 200 & 0.410 & 0.353 & 0.965 & 0.375 & 2656 \\
\runO & \modelB & 200 & 0.415 & 0.387 & 0.870 & 0.385 & 2535 \\
\runP & \modelB & 200 & 0.380 & 0.375 & 0.850 & 0.315 & 2457 \\
\runQ & \modelB & 200 & 0.430 & 0.369 & 0.905 & 0.375 & 2350 \\
\runR & \modelB & 60 & 0.383 & 0.253 & 0.933 & 0.333 & 2759 \\
\runS & \modelB & 60 & 0.373 & 0.290 & 0.915 & 0.305 & 2466 \\
\runT & \modelB & 60 & 0.400 & 0.324 & 0.000 & 0.000 & 2772 \\
\bottomrule
\end{tabular}
}
\end{table}

%% file: paper/tables/ticket_error_summary.tex
\begin{table}[H]
\centering
\caption{Ticket routing error summary (compact; WF = workflow success, p95 in ms).}
\label{tab:ticket-error-summary}
\scriptsize
\setlength{\tabcolsep}{2pt}
\begin{tabular}{l r r r r r}
\toprule
Run & N & Acc & F1 & WF & p95 \\
\midrule
CS prompt v1 (seed 1337) & 200 & 0.325 & 0.280 & 0.215 & 3203 \\
CS prompt v2 (seed 1337) & 200 & 0.305 & 0.269 & 0.165 & 2948 \\
IT prompt v1 (seed 1337) & 200 & 0.410 & 0.359 & 0.390 & 2912 \\
IT prompt v2 (seed 42) & 200 & 0.430 & 0.369 & 0.375 & 2350 \\
\bottomrule
\end{tabular}
\end{table}

%% file: paper/tables/ticket_error_full.tex
\begin{table}[H]
\centering
\caption{Ticket routing error summary (full).}
\label{tab:ticket-error-full}
\footnotesize
\setlength{\tabcolsep}{3pt}
\resizebox{\linewidth}{!}{%
\begin{tabular}{l r r r r r}
\toprule
Run & N & Routing Acc & Macro F1 & Workflow & p95 latency (ms) \\
\midrule
CS prompt v1 (seed 1337) & 200 & 0.325 & 0.280 & 0.215 & 3203 \\
CS prompt v2 (seed 1337) & 200 & 0.305 & 0.269 & 0.165 & 2948 \\
IT prompt v1 (seed 1337) & 200 & 0.410 & 0.359 & 0.390 & 2912 \\
IT prompt v2 (seed 42) & 200 & 0.430 & 0.369 & 0.375 & 2350 \\
\bottomrule
\end{tabular}
}
\end{table}